\documentclass[11pt]{article}
\usepackage[margin=1in]{geometry}
\usepackage{graphicx,amsmath,amssymb,booktabs,hyperref,xcolor}
\usepackage[T1]{fontenc}
\usepackage{natbib}
\graphicspath{{fig/}}
\hypersetup{colorlinks=true,linkcolor=blue,citecolor=blue,urlcolor=blue}

\newcommand{\fpf}{\phi}
\newcommand{\Nstarts}{96}

\title{Prompt--Model Interaction Reaches the Fixed Points\\
{\large A deterministic, task-free structural readout --- and the factorizations of it that failed}}

\author{Nicol\'as Vera Z\'u\~niga\\
Independent Researcher, Chile\\
\texttt{nicovera@quetru.cl}
}
\date{}

\begin{document}
\maketitle

\begin{abstract}
That a prompt's effect on a language model is not a property of the prompt is established: prompts
optimised for one model degrade on another, formats have no model-independent valence, and benchmark
rankings reorder under semantically neutral reformatting. All of that evidence is about \emph{task
accuracy}, and a task readout cannot say where the interaction lives --- in the machinery of
performing tasks, or in the conditional distribution itself. We ask on a readout with no task in it:
the fixed-point structure of the short-window argmax map
$x_{t+1} = \arg\max_x p(x \mid x_{t-1}, x_t)$, censused from $\Nstarts$ starts. It is deterministic,
so nothing can be helped or hurt; and it exists only at short windows --- four of six models lose it
entirely by window $16$ --- so everything here is a statement about how a model reads a
\emph{fragment}. Two results. First, the interaction reaches this readout at full magnitude: nine
tokens of conditioning move the fixed-point fraction across most of its range, change a four-way
structural class, and reorder models, while an instruction-tuning intervention that moves IFEval by
$60.5$ points moves the class by zero. Second, nothing we proposed carries the effect. Prefix length
fails: the effect is not monotone. Four phenomenological factors --- prose-versus-markup, a
universal direction, bidirectionality as a model property, instruct-resistance --- were each
withdrawn within one run of being proposed, dissolved by widening the sample. And the nearest
mechanistic account, attention-sink dominance of early tokens, predicts the sign of the shift on
$2$ of $5$ models --- chance --- with a length-by-content cross showing why: the account holds on
real text and fails on the uniformly random input our probe feeds it, so we operate outside its
regime rather than against it. One fixed nine-token prefix drives four models toward fixed-point
fraction $0$ and two toward $1$; a single beginning-of-sequence token takes one model from $0.21$ to
$0.91$ while collapsing others; the bidirectionality survives in-distribution starts. On this
readout the unit of explanation is the prompt--model pair. We close with the discipline that caught
our own four factors, whose recurring error has a name: a criterion with a shape applied to a
quantity with no room to vary.
\end{abstract}

\section{Introduction}\label{sec:intro}

\subsection{What is already known}\label{sec:intro-known}

Language model evaluations are sensitive to the prompt in ways that are not nuisances. Semantically
neutral reformatting reorders leaderboards by up to eight positions\citep{alzahrani2024benchmarks};
under adversarial prompt selection ``any model can be promoted to first place'' in a six-model,
eleven-dataset embedding benchmark\citep{kostiuk2026oneprompt}; per-prompt performance rankings agree
across models only weakly, at Kendall's $W = 0.238$\citep{cao2024worstprompt}; format performance
``only weakly correlates between models''\citep{sclar2024format}. A prompt optimised for one model is
suboptimal on another --- named \emph{Model Drifting}\citep{wang2025promptbridge}, and anticipated by
earlier work establishing that a good prompt ``binds to the nature of the LLM in
question''\citep{chen2023mapo}. The best prompt templates do not transfer between models, even within
a single model family\citep{voronov2024format}. Adjacent to the axis we vary, though on a different
readout, \citet{mahaut2025repetitions} find that prompt \emph{type} changes the mechanism of
repetition itself: repetition induced by in-context learning setups that explicitly require copying
``relies on a dedicated network of attention heads that progressively specialize over training,
whereas naturally occurring repetition emerges early and lacks a defined circuitry''. That work
measures no fixed points.

We take all of this as given. \textbf{This paper does not claim to discover prompt--model
interaction.}

\subsection{The open question}

Every result above is measured on \emph{task performance}: accuracy, win rate, mean average
precision, benchmark position. Whether the interaction is a fact about how models perform tasks ---
instruction following, format parsing, answer extraction --- or a fact about the conditional
distribution itself, is not settled by that evidence, because a task readout cannot separate the two.

We therefore ask the question on a readout with no task in it. The \emph{short-window argmax map}
iterates $x_{t+1} = \arg\max_x p(x \mid x_{t-1}, x_t)$ deterministically; its fixed-point structure is
a property of the conditional alone. There is no answer to get right, no format to parse, no
verbalizer, and --- the map being deterministic --- no sampling noise. A prefix cannot help or hurt
performance here, because there is no performance.

\subsection{The claim, and the ladder}\label{sec:ladder}

The interaction reaches this readout --- and then resists explanation in a specific, documentable
way: every factorization of the effect we proposed, by a property of the prefix, a property of the
model, or the nearest mechanistic account, failed when the sample widened. The paper is organised as
that ladder.

\begin{enumerate}
\item \textbf{The locus} (\S\ref{sec:moves}). Nine tokens move the fixed-point fraction across most
      of its range and change a four-way structural class, while instruction tuning --- $+60.5$
      IFEval points on a matched pair --- moves the class by zero. The effect is large and
      selective, and it is a fact about the conditional, not about task machinery.
\item \textbf{Prefix-side factors fail} (\S\ref{sec:prefix}). The effect is not monotone in prefix
      length, and both factors we proposed on the prefix's content died on widening
      (Table~\ref{tab:withdrawals}, rows 1 and 3).
\item \textbf{Model-side factors fail} (\S\ref{sec:interaction}). A universal direction,
      bidirectionality as a model property, and instruct-resistance died the same way (rows 2, 4,
      5). What remains is pairwise: one fixed prefix drives four models toward $0$ and
      two toward $1$, and the pattern survives in-distribution starts.
\item \textbf{The mechanistic factor fails to carry the sign} (\S\ref{sec:sink}). Measured on the
      same forward passes, attention-sink strength agrees with the fixed-point shift on $2$ of $5$
      models --- chance --- and a length-by-content cross shows the account holds on real text and
      fails on our probe's input: outside its regime, not contradicted.
\item \textbf{The discipline} (\S\ref{sec:discipline}): what kept the factor search honest, and the
      recurring error it caught in ourselves.
\end{enumerate}

We do not claim that no factorization exists. We claim that five natural ones did not survive
widening, we state the criterion each was held to (\S\ref{sec:conclusion}), and we invite the sixth.

\begin{table}[t]\centering
\small
\begin{tabular}{lll}
\toprule
claim & died on & source \\
\midrule
``prose and markup differ, and the sign flips by model'' & mid-range models & F147 $\to$ F151 \\
``the effect is unidirectional: 18 of 18 arms down''      & a wider model set & F151 $\to$ F152 \\
``no text raises two models''                             & a wider corpus & F153 $\to$ F154 \\
``bidirectionality is a model property''                  & a wider text class & F154 $\to$ F155 \\
``instruct models resist raising''                        & ONE additional text & F156 \\
\bottomrule
\end{tabular}
\caption{The factor record, and the paper's roadmap. Each proposed factor was withdrawn within one
run of being proposed, always because widening the sample dissolved it into an interaction that had
been undersampled. \S\ref{sec:prefix}--\S\ref{sec:sink} walk the ladder; \texttt{findings.md}
carries the banners.}
\label{tab:withdrawals}
\end{table}

\paragraph{Relation to the companion instrument.} The estimator family is the one whose validity
conditions we established in \citet{veraz2026probes}, which asked which readings of an iterated
probe belong to the construction and which to the model, and gave the two-axis test separating them.
In that vocabulary, the prefix is a construction axis; the window sweep in Setup scopes the readout
itself; and this paper's result is that the axis's effect on the readout is model-conditioned in
sign --- the interaction is what survives the discriminator.

\section{Setup}

\paragraph{Estimator.} Iterate the map from $\Nstarts$ random two-token starts and census where
trajectories land. The readout is the \emph{fixed-point fraction} $\fpf$, the proportion of starts
terminating in a fixed point; trajectories are also classified \textsc{funnel} (one dominant
attracting fixed point), \textsc{none} (cycles or wandering), \textsc{fragmented} (many small basins),
\textsc{borderline}. The funnel geometry --- many states feeding one self-continuing token --- was
derived theoretically by \citet{fu2021repetition}, whose inflow analysis explains why trajectories
concentrate; the classes here are that geometry measured on a model's own conditional rather than on
corpus counts. Two rows of Table~\ref{tab:moves} use the \emph{share}, this readout's
stochastic sibling from the same instrument family --- the fraction of settled mass in the dominant
attractor of the sampled ring \citep{veraz2026probes}; everything else in the paper is the
deterministic census above.

\paragraph{Domain.} A token prefix prepended to the context before every forward pass, implemented as
a wrapper around the model rather than a change to the estimator, so the measured quantity is
provably unchanged. Kinds: \textsc{raw} (none), \textsc{bos}, \textsc{text} (ordinary text truncated
to a stated count), \textsc{template} (the model's own chat template), \textsc{struct} (high markup
density, Table~\ref{tab:withdrawals}).

\paragraph{Scope: a short-window estimator, by nature and not by choice.}
Generalising the map so the state is the last $W$ tokens and sweeping $W \in \{2,4,8,16\}$, the
\emph{readout itself} disappears: raw $\fpf$ falls from $0.22$--$0.70$ at $W{=}2$ to $0.000$ on four
of six models by $W{=}16$, and to $\leq 0.10$ on a fifth. Only one model retains measurable
fixed-point structure at the longest window. The fixed-point structure of the argmax map is therefore
a short-window property: widen the window and there is nothing left to measure, so questions about the
domain at $W{=}16$ are not answered in the negative --- they are unaskable.

Everything below is accordingly a statement about a model reading a \emph{fragment}. We put this in
scope rather than in limits because it is a property of the construction, not a shortfall of the
evidence; \S\ref{sec:limits} does not weaken it.

\paragraph{Tolerance.} $\fpf$ is quantised at $1/\Nstarts \approx 0.0104$. A shift counts only if it
exceeds that model's own tolerance $\max(4/\Nstarts,\, 2s)$, with $s$ the across-seed range
on the arm in question. \S\ref{sec:discipline} explains why this is stated so prominently.

\section{The locus: the interaction reaches a task-free structural readout}\label{sec:moves}

\begin{table}[t]\centering
\small
\begin{tabular}{llll}
\toprule
readout & domain change & effect & source \\
\midrule
structural CLASS   & 9-token chat template & $\fpf: 0.948 \to 0.000$, class changes & F144 \\
class, \textsc{fragmented} model & every domain tested & changes under all of them & F151 \\
model RANKING      & chat template & $\rho = +0.400 / +0.300$ vs a $0.6$ gate & F145 \\
scalar VALUE       & chat scaffold & $0.1327$ shift vs a $0.0406$ gate ($3.3\times$) & F135 \\
$\fpf$             & 9 tokens of prose & $0.714 \to 0.005$ & F151 \\
\bottomrule
\end{tabular}
\caption{Conditioning moves a task-free structural readout at magnitudes comparable to those the
literature reports for accuracy. Rows 3--4 are measured on the share, the stochastic sibling readout
(Setup); the ranking row is an instance of known ranking instability (\S\ref{sec:intro-known}); the
class rows have no accuracy analogue.}
\label{tab:moves}
\end{table}

\paragraph{A contrast that bounds the interpretation.} Instruction tuning moves IFEval by $+60.5$
points on one matched pair and moves the structural class by \emph{zero}.
This readout is therefore not sensitive to everything: a large behavioural intervention leaves it
where it was, while nine tokens of conditioning move it completely. Whatever the domain does here is
not generic fragility.

The remainder of the paper asks what carries the effect, and reports the failure of each candidate
in turn.

\section{Prefix-side factors fail}\label{sec:prefix}

\paragraph{Length is not the factor.}
Sweeping \textsc{raw} (0 tokens) $\to$ \textsc{bos} (1) $\to$ \textsc{text} (matched to that model's
own template length) $\to$ \textsc{template}, three of the four models with enough span to carry a
shape are \textbf{not} monotone in length. One runs $0.948 \to 0.005 \to 1.000 \to 0.000$: a single
BOS token nearly annihilating its structure, nine tokens of prose restoring it to a perfect funnel,
nine tokens of chat markup destroying it again.

The length-matched text arm is what makes this readable --- same token count as the model's own
template, none of its structure --- so \textsc{text} versus \textsc{template} isolates kind at fixed
length. The accuracy-level prompt space has the same shape: it is reported as ``largely
non-monotonic'', with graded atomic format changes giving monotone accuracy triples $32.4\%$ and
$33.6\%$ of the time against a $33.3\%$ chance rate\citep{sclar2024format}. Ours is that shape on a
readout with no task, no format and no examples in it.

\paragraph{The prefix's content is not the factor either.} Two content factors followed the length
result and died on widening (Table~\ref{tab:withdrawals}, rows 1 and 3): ``prose and markup differ,
and the sign flips by model'' did not survive the mid-range models (F147 $\to$ F151), and ``no text
raises two models'' did not survive a wider corpus (F153 $\to$ F154). Their narratives continue in
\S\ref{sec:interaction}, because what dissolved them was in each case the \emph{model} side of the
sample.

\section{Model-side factors fail, and what remains is pairwise}\label{sec:interaction}

\paragraph{A universal from $n=2$.} On two mid-range instruct models all $18$ domain arms moved down,
and every apparent up-shift to that point proved mechanical --- one saturating exactly at the ceiling,
one inside its own seed spread, one starting from the floor. We wrote ``unidirectional, without
exception''. Seven base models produced two bidirectional counterexamples immediately.

\paragraph{A null we wanted.} Having found that high-markup text raises some models more often than
prose, we noted two instruct models resisting every text tried, and pre-registered that a null would
be \emph{the first factor here to survive a widening} --- a positive claim this paper otherwise lacks.
One text of twelve refused it: a C source-file header takes \texttt{gemma-2-2b-it} from $0.714$ to
$0.917$. We report the null we wanted and did not get.

\paragraph{What survives, unpromoted.} Instruct models were raised on $1$ of $24$ text-model units
against $11$ of $48$ for base models on the \emph{same} twelve texts. Two instruct clusters cannot
test that, and Table~\ref{tab:withdrawals} is four consecutive instances of this kind of gap
dissolving. We record it and decline to promote it.

\paragraph{What remains when the factors are gone.} With no prefix-side and no model-side factor
surviving, the residual is the pair itself.

\begin{table}[t]\centering
\small
\begin{tabular}{lrrl}
\toprule
model & raw $\fpf$ & under the prefix & direction \\
\midrule
Falcon3-1B-Base       & 0.214 & \textbf{0.979} & up \\
Minerva-3B-base       & 0.328 & \textbf{1.000} & up \\
pythia-410m-deduped   & 0.427 & 0.016 & down \\
Qwen1.5-1.8B          & 0.510 & 0.000 & down \\
SmolLM-1.7B           & 0.562 & 0.000 & down \\
Qwen2.5-1.5B-Instruct & 0.573 & 0.005 & down \\
\bottomrule
\end{tabular}
\caption{One fixed nine-token prefix (a table-of-contents fragment) applied to six models
screened to have headroom in both directions. The prefix is the same source text at the same index,
truncated to nine tokens under \emph{each model's own} tokenizer --- so the length is held fixed and
the character span varies slightly between models; per-cell token counts are in the results file. The bidirectionality survives the in-distribution
control of \S\ref{sec:limits} (F160). Source: F154,
\texttt{results/text\_interaction.json}.}
\label{tab:p1}
\end{table}

Table~\ref{tab:p1} is the structural-readout instance of a known accuracy-level phenomenon. We claim
no priority for the pattern. That prompt quality does not certify across weights is in the abstract
of \citet{lu2022ordered} --- ``a given good permutation for one model is not transferable to
another'', with prompt rankings correlating at $0.05$ between their $2.7$B and $175$B models --- and
one fixed permutation there already moves accuracy from $88.7\%$ to $51.6\%$ between two sizes of
GPT-2. The phenomenon is named \emph{Model Drifting} by \citet{wang2025promptbridge}, and MAPO
established prompt effectiveness as model-specific\citep{chen2023mapo}.
What Table~\ref{tab:p1} adds is that it holds where there is nothing to be good or bad at: the same
prefix is not merely \emph{less useful} to some models, it moves a deterministic property of their
conditional in opposite directions. The same battery contains the converse exhibit: under a
second fixed prefix from the same source, \texttt{Qwen1.5-1.8B} --- a faller in
Table~\ref{tab:p1} --- rises to $0.979$ while \texttt{SmolLM-1.7B} stays at
$0.000$.
The direction is set by the pair, not by the model.

The nearest prior work on a structural readout uses iterated text-level transmission chains under
stochastic sampling ($T = 0.8$, top-$p$ $0.95$) over six instruction-tuned models, and reports that
``different models lead the distributions to be shifted in opposite directions'' on a scalar text
property --- text length shortens under GPT3.5 and Llama3-8B and lengthens under Mixtral-8x7B and
GPT-4o-mini\citep{telephone}. A separate line reports iterated-map attractor states as
\emph{model-independent}\citep{paraphrase2cycle}; \S\ref{sec:limits} takes up why both claims can
stand, each in its own scope.

\section{The mechanistic factor: where the sink account predicts magnitude, we observe
sign}\label{sec:sink}

This is the finding we cannot place in the existing literature, and the one we would most like
refuted.

\paragraph{What the sink account says.} Initial tokens dominate: their presence or absence moves a
scalar readout by orders of magnitude, holding weights and decoding
fixed\citep{xiao2023streamingllm}. The proposed mechanism is \emph{positional}, not semantic ---
substituting the first four tokens with the linebreak token restores perplexity nearly as well as the
originals ($5.60$ against $5.40$ on Llama-2-13B, versus $5158.07$ with those positions dropped),
because SoftMax normalisation forces attention mass onto early tokens regardless of content. On that
account the sink ``acts more like key biases, storing extra attention scores, which could be
non-informative and not contribute to the value computation''\citep{gu2025sink}, and the reported
cross-model variation is in \emph{magnitude and saturation point}, not in sign: adding initial tokens
moves the readout the same way in every model examined.

\paragraph{What we observe.} A single BOS token takes \texttt{Falcon3-1B-Base} from $0.214$ to
$0.906$,
and collapses other models toward zero on the same arm. The sign is not shared.

\paragraph{We measured the mechanism, not just our own readout.} Comparing a BOS-prefixed context
against a \emph{length-matched} ordinary one --- so the arms differ only in the identity of position
0 --- sink strength and $\fpf$ move the same way on 2 of 5 models, which is chance
(Table~\ref{tab:sink}).
Sink strength does not predict the sign of the fixed-point shift.

\begin{table}[t]\centering
\small
\begin{tabular}{lrrl}
\toprule
model & $\Delta$sink & $\Delta\fpf$ & agree \\
\midrule
Falcon3-1B-Base       & $+0.0403$ & $+0.693$ & yes \\
Minerva-3B-base       & $+0.0675$ & $-0.323$ & no  \\
pythia-410m-deduped   & $-0.0197$ & $-0.422$ & yes \\
Qwen1.5-1.8B          & $+0.0298$ & $-0.510$ & no  \\
SmolLM-1.7B           & $+0.1484$ & $-0.562$ & no  \\
\bottomrule
\end{tabular}
\caption{One BOS token, against a length-matched ordinary prefix. The mechanism and the structural
consequence agree on 2 of 5 models. Source: F158, \texttt{results/sink\_vs\_fixedpoint.json}.}
\label{tab:sink}
\end{table}

\paragraph{The regime differs on CONTENT, not length --- and this narrows our claim.}
The obvious explanation for the mismatch is context size, and it is wrong. Crossing five lengths
($2$--$512$) with two contents (uniformly random tokens versus real text) on the same six models: sink
\emph{concentration} --- attention to position $0$ times sequence length --- rises with length on
$6$ of $6$ models in both contents, from ${\sim}2\times$ uniform at $n{=}2$ to ${\sim}144\times$ at
$n{=}512$, so the measurement reproduces the phenomenon. But on \textbf{real text no resolved model
shows sink decreasing under BOS at any length $\geq 8$}, while on \textbf{random tokens some do at
every length}.

The attention-sink account therefore holds in its own regime. Our probe draws uniformly random
two-token starts, which is out-of-distribution input in a way real text is not, and that --- rather
than context size --- is where the account's uniformity fails. We accordingly do \emph{not} claim to
contradict it. The claim is narrower: \textbf{the probe operates outside the regime its nearest
mechanistic account describes}, so that account cannot be assumed to explain what the probe reports,
and the failure above of sink strength to predict the sign is unexplained rather than
contradictory.

\paragraph{Why this is a real tension and not a definitional one.} A positional, content-free
mechanism predicts a shared direction: if the first token's role is to absorb attention mass, then
supplying one should push every model the same way, differing only in how much. Our readout is
structural rather than perplexity-like, so the two do not measure the same quantity --- but a
mechanism advertised as content-independent and architecture-general should not reverse sign across
models on \emph{any} downstream property, and Table~\ref{tab:sink} shows the mechanism's own
magnitude failing to carry the sign even where we measure both at once.

\section{The discipline}\label{sec:discipline}

\paragraph{A criterion with a shape must not be applied to a quantity with no room to vary.} Our most
frequent error by a wide margin. Instances: flat sequences scored as monotone-or-not when their whole
span was one census start; one trajectory in $\Nstarts$ flipping a monotonicity verdict; two floored
arms counted as ``agreement'' when neither could disagree; a floor/ceiling baseline scoring
$9/9 = 100\%$ where no unit could have contradicted it; and twice the defect appearing \emph{inside}
the anti-vacuity gate written to prevent it --- once excluding the most decisive model in a run while
admitting one whose entire effect was half a census start.

\paragraph{Its sibling.} A confident universal from $n=2$ is not vacuous in the same way --- a
counterexample was possible --- but unlikely enough that the claim was never at risk. Both are cheap
to catch by widening before claiming.

\paragraph{Protocol.} (i) Screen for headroom before testing direction; (ii) gate each question
separately, since ``is the contrast robust'' needs the gap real while ``is the value stable'' needs
the values off the floor; (iii) judge each shift against its own noise, not a pooled tolerance;
(iv) verify anti-vacuity rather than assume it, even when models were selected for it; (v) refuse
tests that cannot fail informatively at the available $n$, and record the refusal before seeing the
numbers.

\paragraph{Pooling.} High-markup text raised two of three models more often than prose and the pooled
ratio ($7/18$ vs $3/18$) cleared our pre-registered bar --- but the third model reversed. A
consistency gate added before the verdict refused to read the pooled number, on the grounds that
pooling across units pointing different ways is a Simpson's shape. Without it we would have reported
a predictive rule.

\section{Limits}\label{sec:limits}

\paragraph{We do not claim the concept.} Prompt--model interaction, prompt non-transferability and
ranking instability under formatting are prior art (\S\ref{sec:intro-known}). Our contribution is that
the interaction reaches a deterministic, task-free structural readout, that no factor we tested
carries it, and the sign behaviour of \S\ref{sec:sink}.

\paragraph{A neighbouring result that reads as a contradiction, and is a scope collision.}
\citet{paraphrase2cycle} report iterated-map attractor states as model-\emph{independent}: ``the
attractor states are not confined to a single model's parameter space. Instead, they reflect a
more general statistical optimum that multiple LLMs gravitate toward.'' Read beside
Table~\ref{tab:p1}, the two claims point opposite ways. The two measurements, however, share no
object: their iteration is text-level, through an instruction-conditioned paraphrase call under
stochastic decoding, where ours is a token-level deterministic argmax map; their attractor is a
$2$-period edit-distance cycle, not a fixed point; and their model-independence is inferred from
\emph{alternating} models within one chain plus a cross-model perplexity check, not from one
prefix held fixed across separate models --- the design Table~\ref{tab:p1} runs. Neither dataset
therefore contradicts the other. What collides is the phrasing: their conclusion is stated as a
property of LLMs at large, our regime is a counterexample to that phrasing at large, and the
symmetric correction is to keep both claims scoped, which we do.

One reading is consistent with both results, and is testable: their loop contains a task --- an
instruction the chain obeys at every step --- and ours was built to contain none. If the shared
attractor belongs to the task layer that instruction-tuned models hold in common,
model-independence there and model-conditioned sign here are the two sides of the distinction
this paper draws between task machinery and the conditional. Their protocol with the instruction
varied, or run over base models, would test it; nothing in our data does.

\paragraph{Small cohorts, one estimator family.} The largest single comparison is six models and twelve
texts, at one prefix length, from three English sources, under one operationalisation of
``structural''. Table~\ref{tab:withdrawals} is why boundaries here read ``these models resisted these
texts'' rather than generally: four times, the general form was wrong.

The window sweep bounds the estimator rather than the cohort. At $W{=}4$ the domain effect has already
decayed on five of six models --- though one, \texttt{Qwen2.5-1.5B-Instruct}, shows it undiminished
($\lvert\Delta\fpf\rvert = 0.547$ at $W{=}4$ against $0.521$ at $W{=}2$) before its structure
collapses by $W{=}8$. One model is not a rate and we do not promote it; it is recorded because a
reader will reasonably ask whether the two-token result is knife-edge, and on that model it is not.

\paragraph{Confounds we could not break.} \texttt{gemma-2-2b-it} is the only \textsc{fragmented} model
in the cohort, so class and cohort are confounded in it. \texttt{Falcon3-3B-Instruct} is excluded from
direction work: its raw census varies $0.615$--$0.792$ across four seeds, so its tolerance swamps
every domain shift it has.

\paragraph{How the prior art here was checked.} The novelty gate that produced this framing refuted
$13$ of $74$ extracted claims for overreaching their own sources,
so its summaries were not citable. Every work cited above was therefore read at the source --- the
abstract where the claim is in the abstract, the full text where it is not --- and the supporting
quote recorded beside its key in \texttt{CITATIONS.md}.

\paragraph{The probe's own input distribution, tested.}
Our census draws starts uniformly from the vocabulary --- token pairs no corpus contains --- and
\S\ref{sec:sink} shows the nearest mechanistic account behaves differently on exactly such input.
That raises the obvious worry that the domain effects are an out-of-distribution artefact, and it is
not hypothetical: earlier work of ours on a sibling readout found precisely that, reading a collapse
under one BOS token as the signature of an OOD prompt rather than of model dynamics.

Holding everything fixed but the start distribution --- uniform token pairs versus \emph{adjacent}
pairs drawn from real text --- the domain effect survives on $6$ of $6$ models, keeping its sign and
rough magnitude, \textbf{including the bidirectionality of Table~\ref{tab:p1}}: one model still rises
to $0.964$ while the rest collapse. The effect is therefore not an artefact of seeding the loop with
nonsense.

Two cautions belong with that. Four of the six reach $\fpf \approx 0$ under BOS in \emph{both}
regimes, so their $\Delta\fpf$ is floor-bounded and its magnitude cannot be compared across regimes
--- a difference there measures the starting point, not the effect. And baseline $\fpf$ itself shifts
by $-0.271$ on one model, so this is a comparison between two regimes rather than a robustness check
within one.

\paragraph{A gap this still cannot close.} $\fpf$ remains unmeasurable at long context, since the
census is defined at two-token starts. We can bound where the sink account applies but cannot connect
it to our readout in the regime where that account holds.

\paragraph{Note added.} \S\ref{sec:conclusion} invites a sixth factorization. Between freezing this
draft and submitting it one was attempted, and held to the criterion stated there. The candidate
carrier was the endpoint token: that a prefix selects which token trajectories are pulled toward,
while the model decides whether that token self-continues --- two vectors whose product would fix a
sign where no single factor had. It faced a widening that had been pre-registered on coverage grounds
\emph{before} the candidate existed, which is the ordering the criterion requires. The strong form
died on one cell: a model rises to $\fpf = 0.995$ under a prefix whose margin for the predicted token
never turns positive.
What survives is not a factor but another interaction. Across the nine models carrying each arm, the
best cross-model agreement on an arm's modal endpoint token is $4$ of $9$,
so which token is selected is itself a model--prefix interaction rather than a property of the
prefix. The repository's findings ledger carries the full sequence (F162--F166); the paper's claims
stand as frozen.

\section{Conclusion}\label{sec:conclusion}

Prompt--model interaction is established for task performance. We show it reaches a property with no
task in it --- the fixed-point structure of a model's own deterministic argmax map, a short-window
readout of how a model treats a fragment --- and that on this readout it resisted every factorization
we could propose. Prefix length is not the factor: the effect is non-monotone. The prefix's content
is not the factor: both content factors died on widening. The model alone is not the factor: a
universal direction, bidirectionality as a model property and instruct-resistance died the same way,
and one fixed prefix drives four models toward zero and two toward one, surviving in-distribution
starts. The nearest mechanistic account is not the factor: its own quantity, measured on the same
forward passes, carries the sign at chance, and the account holds on real text while failing on the
probe's input --- outside its regime, not contradicted.

The criterion each factor was held to is the one we offer forward: \emph{a factor survives if a
pre-registered widening it did not choose fails to dissolve it}. Five did not survive. Until one
does, the unit of explanation at this readout is the prompt--model pair, and the discipline that kept
the search honest --- refuse any criterion with a shape wherever the quantity has no room to vary ---
is what we would most want carried forward.

\bibliographystyle{plainnat}
\bibliography{refs}

\end{document}